Artificial Intelligence in Surgery

# Neural Networks and Deep Learning

Deepak Alapatt[1†], MSc, Pietro Mascagni[1, 2†], MD, Vinkle Srivastav[1], MS, Nicolas Padoy[1], PhD

1. ICube, University of Strasbourg, CNRS, IHU Strasbourg, France

2. Fondazione Policlinico Universitario Agostino Gemelli IRCCS, Rome, Italy

Deep neural networks power most recent successes of artificial intelligence, spanning from self-driving cars to computer aided diagnosis in radiology and pathology. The high-stake data intensive process of surgery could highly benefit from such computational methods. However, surgeons and computer scientists should partner to develop and assess deep learning applications of value to patients and healthcare systems. This chapter and the accompanying hands-on material were designed for surgeons willing to understand the intuitions behind neural networks, become familiar with deep learning concepts and tasks, grasp what implementing a deep learning model in surgery means, and finally appreciate the specific challenges and limitations of deep neural networks in surgery. For the associated hands-on material, please see https://github.com/CAMMA-public/ai4surgery.

**HIGHLIGHTS**

o   The growing availability of digital data and computational power enables the training of deep neural networks useful for real world applications.

o   Deep neural networks are typically composed of multiple convolutional layers used to efficiently extract information from high dimensional inputs, pooling layers to reduce dimensions, and fully connected layers to aggregate neuron activations into output values.

o   Neural networks learn to approximate a function by forward propagating the input layer by layer, calculating a loss comparing the current output to the ground truth and then updating the network's weights and biases through backpropagation.

o   Deep architectures are selected based on the nature of the input and desired output data to perform tasks such as classification, detection, semantic segmentation and temporal recognition.

o   Deep learning strategies to guarantee the security of sensitive medical data, train networks using less supervision, increase model explainability and deliver real-time predictions in the OR are being developed to generate value in surgery.

# TABLE OF CONTENT

## INTRODUCTION

*"Instead of trying to produce a programme to simulate the adult mind, why not rather try to produce one which simulates the child's? If this were then subjected to an appropriate course of education one would obtain the adult brain."*

Alan Turing, 1950, I. Computing Machinery and Intelligence

The question of whether machines could be capable of thinking and learning like humans has always allured mankind. As briefly discussed in Chapter 1 (A Brief History of AI), ancient mythology dating back further than 2500 years ago already makes references to modern concepts such as self-moving objects, robots and what we now refer to as artificial intelligence (AI). The recent surge in deep learning is contributing towards turning these antic fantasies into today's reality. Deep learning breakthroughs in fields ranging from image and speech recognition to game playing have been widely covered by the press, generating enthusiasm in the general public as well as among businesses and funding agencies. Replicating biological neurons on silica chips is by no means a novel idea; however, the incredible amount of digital data we now ubiquitously generate – together with the growing availability and decreasing cost of computational power – make training deep neural networks practical. Furthermore, open source programming frameworks have lowered the barrier to entry, allowing quick prototyping for real-world applications.

Deep learning models are already in use for applications such as automatic recommendations of media content and smart assistants in our homes and self-driving cars. Healthcare is a very active field of research for AI given its social and economic relevance. In this sector, surgery is a particularly interesting and challenging subfield since it is a high-stakes discipline where multiple people interact, make quick decisions based on a large amount of sparse information, and act to alter a patient's anatomy. Despite clear opportunities and widespread hype, deep learning has yet to impact surgical patients. It is thus the ideal moment for surgeons to understand the intuitions behind neural networks, become familiar with deep learning concepts and tasks, grasp what implementing a deep learning model in surgery means, and finally appreciate the specific challenges and limitations to address for deep learning and AI to benefit patients and surgical practices.

## A HISTORICAL PERSPECTIVE

The intuitions behind artificial neural networks were profoundly inspired by those gained studying their biological counterpart, as is often the case with technological innovation. Scientific observations on the human brain date back to the start of the 20[th] century, when Santiago Ramón y Cajal and Camillo Golgi first hypothesized that neurons serve as the fundamental building block of the nervous system, a complex network that extends from the brain to the body periphery. A simplified mathematical formulation of the behavior of a neuron (Warren McCulloch and Walter Pitts, 1943[1]) and the concept that synaptic plasticity contributes to learning (Donald Hebb, 1949[2]) later paved the way to neuropsychology and neural networks. The hierarchical processing of neural networks, an extremely relevant insight linking biological and artificial neural networks, is a concept introduced by Hubel and Wiesel in 1962[3]. They observed that specific neurons in the primary visual cortex of cats responded based on the orientation of edges shown to them in images. They hypothesized that neurons in early layers (i.e. closer to the retina) extract low-level features like edges while down-stream layers extract higher level semantic information.

As the theoretical understanding of the neuron was gaining traction in neuroscience, computer scientists also started prototyping neurons on chips. In 1958, Frank Rosenblatt proposed the first computational model of a neuron known as the *perceptron*. This model takes the weighted sum of inputs and outputs a 1 if the sum exceeds a given threshold, similarly to neurons firing an action potential if a certain threshold potential is reached. Similar to biological neural networks, perceptron's internal parameters update much like neurons' connections strengthen or weaken with learning. Soon after, Bernard Widrow and Marcian Hoff proposed a multi-layer artificial neural network called MADALINE but it was proven to be too computationally expensive given the computers available at the time[4]. This was followed by the "AI winter", a period of substantial decrease in neural network research. Almost 3 decades passed by, until in 1998 Yann LeCun and colleagues demonstrated that their multi-layer *convolutional neural network* (CNN), also called *LeNet[5]*, could accurately recognize handwritten digits, the "Hello world!" of AI. Still, at the time neural networks were trained and tested on small datasets and with limited compute power, limiting their applicability outside of controlled environments. In 2012, Alex Krizhevsky and colleagues trained an extension of LeNet, AlexNet[6], on a large scale image classification dataset, ImageNet[7], using graphics processing units (GPU). AlexNet represented a pivotal point in neural network research. Since then, research on neural networks flourished, leading to improved performance and new applications. For instance, neural networks such as Deeplab[8] or Mask R-CNN[9] can localize different objects in images and Generative Adversarial Networks[10] can generate realistic images starting from completely unrelated data. Research on neural networks is also gaining momentum in the medical field with applications ranging from predicting outcomes using electronic health records[11] to medical imaging analysis[12] and surgical data science[13]. For example, variants of Unet[14] have been used to highlight anatomical structures down to the pixel-level and EndoNet[15], an extension of Alexnet, has been used to analyze surgical workflow in endoscopic videos.

## ARTIFICIAL AND CONVOLUTIONAL NEURAL NETWORKS

A function in mathematics is a relation that assigns one or more unique output values to any given input. It may be trivial to define a function to detect atypical values in routine blood tests; however, defining a function to detect abnormalities in radiological scans may not be as straightforward a task due to the varied appearance of both normal and abnormal scans. Neural networks offer a means to approximate an unknown function when the output values for a large number of sufficiently informative inputs are known but the relationships between them cannot be deduced easily using conventional approaches. This is why neural networks, also known as universal function approximators, are an extremely useful tool for a variety of applications.

### Artificial Neuron and Neural Networks

The fundamental building block of neural networks is the *artificial neuron* (Figure 4-1) or node (hereinafter referred to as neuron) that is used to map a multidimensional input $(x_1, x_2, \ldots, x_n)$ to a single output value.

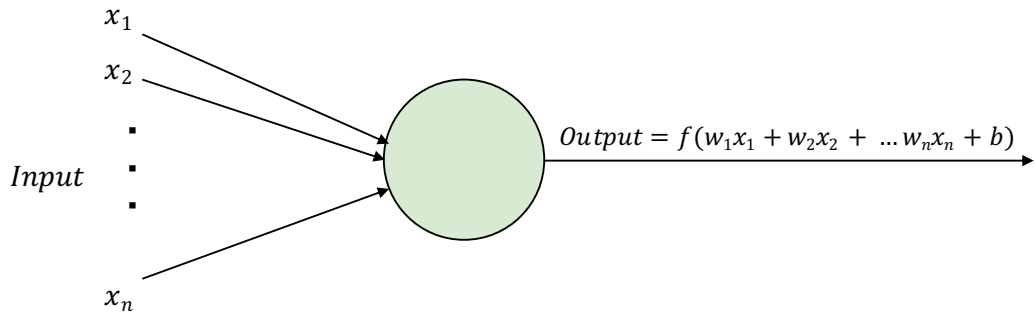

Figure 4-1 Artificial neuron.

The three main components of the neuron are a set of weights $(w_1, w_2, \ldots, w_n)$, a bias ($b$) and an *activation function* ($f$). The weights and biases of the neuron are numeric values or *parameters* that can be used to represent any linear combination of the inputs in the form:

$$w_1 x_1 + w_2 x_2 + \cdots + w_n x_n + b. \qquad \text{(Eq 4-1)}$$

The activation function $f$ then transforms the sum of weighted inputs by applying nonlinearity to the output of that neuron. This operation is performed since these neurons are used to learn real-world representations, which are seldom linear. Some useful activation functions are described in table 4-1. The choice of activation functions is based on a number of factors such as speed of computation, the purpose of the neuron and the nature of the data it operates on. Some of the activation functions mentioned above can be used to "squeeze" the output of the neuron to a fixed range. For instance, *sigmoid* transforms the input function to an output between 0 and 1 (see table 4-1). This type of activation function is known as a *saturating function* and may be useful if the expected output of the neuron is within a fixed and known range of values, for example, to represent the probability of an event occurring where the only plausible values would be between 0 and 1. Rectified Linear Unit (or ReLU) is another commonly used choice of activation function as it is computationally inexpensive to calculate.

| | | |
|---|---|---|
| Step Function | $f(x) = \begin{cases} 1, & x \geq threshold \\ 0, & x < threshold \end{cases}$ | 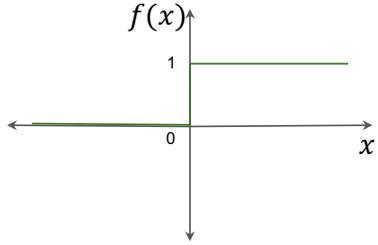 |
| ReLU | $f(x) = \max(0, x)$ | 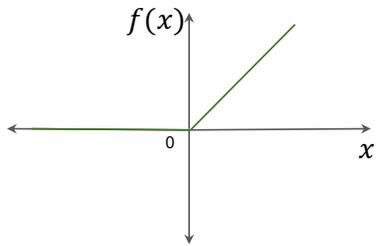 |
| Sigmoid | $f(x) = \dfrac{1}{1 + e^{-x}}$ | 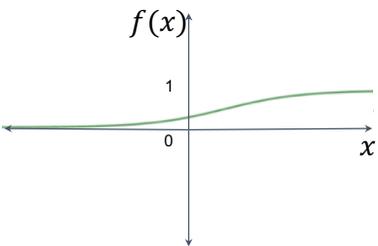 |
| Tanh | $f(x) = tanh(x)$ | 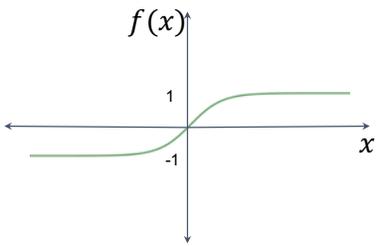 |

Table 4-1 Commonly used activation functions along with their mathematical formulations and graphical representations. *ReLU* and step show a simple nonlinearity at x=0 while sigmoid and tanh represent more complex nonlinear functions.

A *neural network* is a collection of interconnected neurons. Each neuron takes a set of inputs, processes them and sends a signal (a numerical one in this case) that in turn serves as inputs to other neurons. Through this section, we will introduce the key concepts involved in designing and implementing a neural network.

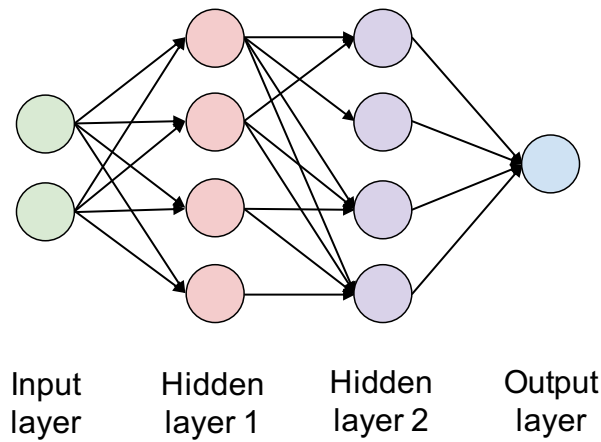

Figure 4-2 Example of a neural network.

In Figure 4-2, each of the different colors represents one layer of the neural network and each of the individual circles, a neuron in that layer. The network consists of an input layer, one or more hidden layers that are used to extract meaningful information from the input and finally an output layer that aggregates this information into a desired form. If the input layer is, for example, a CT scan slice showing a tumor, then the output could represent the probability of malignancy. When there are multiple hidden layers between the input and output layers, the network is said to be a *deep neural network* and training such a network to perform a task is referred to as *deep learning*. Lower layers can be thought to extract simple information, such as edges for images, and higher layers build on that information to develop an understanding of more complex concepts, such as parts and then more complex shapes[16].

Let us now look at how neurons function together in a layer beginning with the simplest kind, the *fully connected layer*. A fully connected layer is basically a collection of neurons each connected to every neuron of the previous layer. Every neuron of the fully connected layer thus receives $n$ input values $(i_1, i_2, \ldots, i_n)$, where $n$ is equal to the number of neurons in the previous layer, and the output of that neuron is given by:

$$w_1 i_1 + w_2 i_2 + \cdots + w_n i_n + b. \qquad \text{(Eq 4-2)}$$

In fact, the output of an entire fully connected layer containing $m$ neurons can be calculated using a single matrix multiplication:

$$\begin{bmatrix} o_1 \\ o_2 \\ \ldots \\ o_m \end{bmatrix} = \begin{bmatrix} w_{11} & w_{12} & \ldots & w_{1n} \\ w_{21} & w_{22} & \ldots & w_{2n} \\ \ldots & \ldots & \ldots & \ldots \\ w_{m1} & w_{m2} & \ldots & w_{mn} \end{bmatrix} \begin{bmatrix} i_1 \\ i_2 \\ \ldots \\ i_n \end{bmatrix} + \begin{bmatrix} b_1 \\ b_2 \\ \ldots \\ b_m \end{bmatrix}. \qquad \text{(Eq 4-3)}$$

Here $o_i$, $(w_{i1}, w_{i2}, \ldots, w_{in})$ and $b_i$ correspond to the output, weights and bias of the $i^{\text{th}}$ neuron in the layer, respectively. A non-linear activation function can then be applied on each of these outputs before passing it as the input to the next layer if any.

Taking the previous example, suppose the CT scan slice showing a tumor is provided as input to a neural network that uses two or more fully connected layers that outputs the probability of malignancy. The slice is first converted into a machine interpretable format as a 2-dimensional (2D) matrix of numeric values representing the brightness of each pixel in the scan slice. In order to be fed to the first fully connected layer of the network, this 2D matrix has to first be transformed into a 1D *vector* (see Figure 4-3a) in order to perform the matrix multiplication described in Eq 4-3. Then the network is trained to find appropriate values for the parameters, i.e. the weights and biases of all the neurons in the network. Finally, the result is computed from different pathways of neurons that extract the visual cues representative of malignancy in the input.

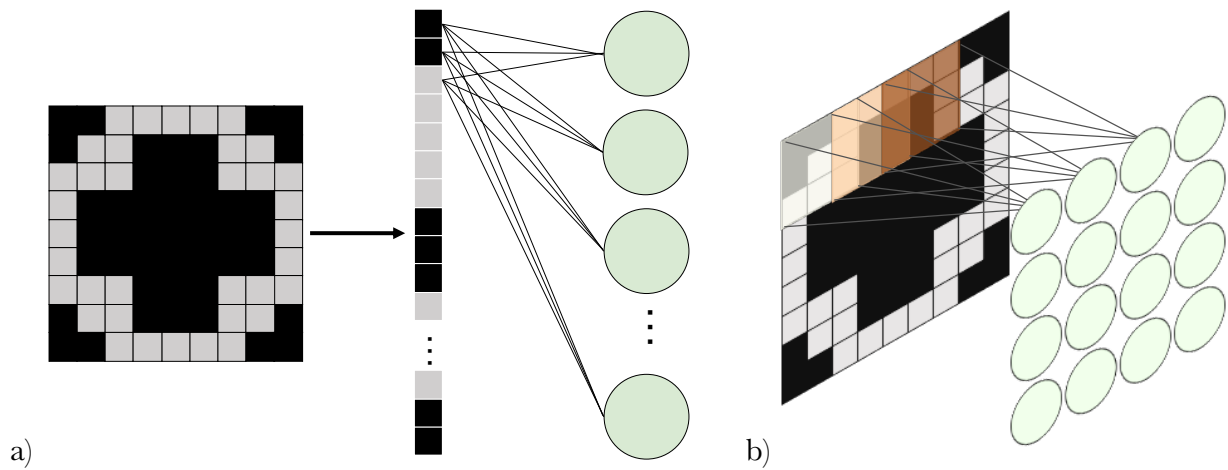

Figure 4-3 Depiction of neurons and their inputs for (a) a fully connected layer and (b) a convolutional layer.

Using only fully connected layers in a network presents two potential issues that can make the process of finding the right parameter values difficult. The first issue is that such a layer does not leverage the information provided by the order in which pixels appear in the original image. For instance, even neighboring pixels in the original 2D image can appear far apart in the transformed 1D vector used as input, as seen in Figure 4-3a. The other issue is that fully connected layers require a large number of parameters to operate on high dimensional inputs, often in the range of millions for a single layer to operate directly on a high-quality image. One way to address both of these issues is to use a 2D layout of neurons that each only operates on a smaller region of the input and that share the same parameter values in order to identify the same visual cues at different positions (Figure 4-3b). Mathematics conveniently packages this solution in the form of the *convolutional layer*, wherein a smaller set of weights and a bias can be "slid" across different parts of the input.

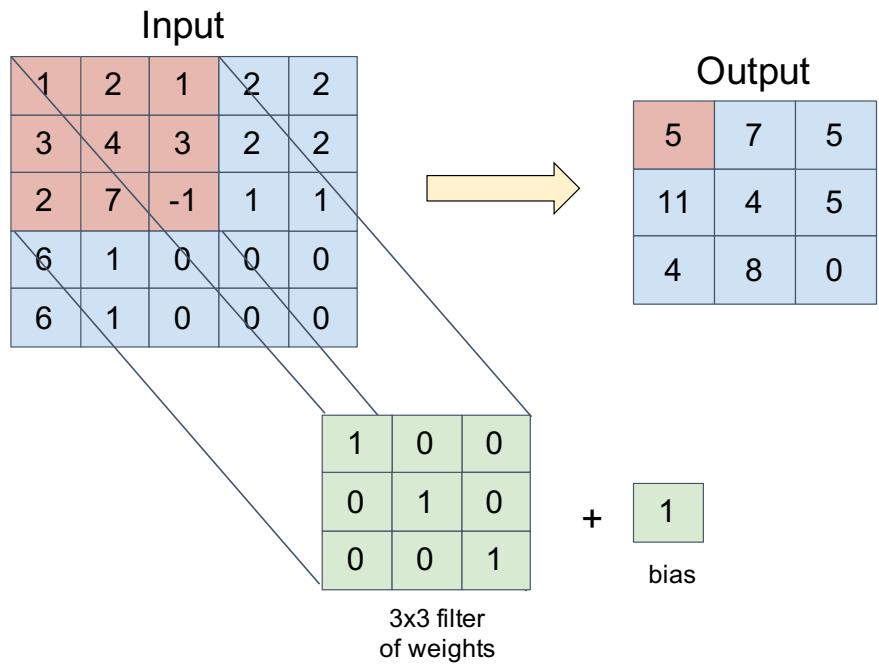

Figure 4-4 Visualization of a 3x3 convolutional filter operating on a 5x5 sized input producing an output of size 3x3.

Figure 4-4 shows a depiction of how the convolution operation works on an input of size 5x5. A small matrix of 3x3 weight values (known as a *filter*) and one bias value is applied on 3x3 patches of the input column-by-column and row-by-row. Because convolution uses the same parameters at different positions, this operation is much more memory efficient than the fully connected layer. In fact, in the given example the convolutional layer only stores 9 weights + 1 bias per neuron whereas a fully connected layer would have required the storage of 25 weights + 1 bias per neuron to operate on the same input. Since each convolutional operation generates a 2D output, using

stacks of such filters would generate a 3D output. When working with 3-dimensional (3D) inputs to convolutional layers, each convolution filter must also be 3D in order to apply it on 3D patches when moving across the length and breadth of the image. The *stride* of the convolution is the amount the filter is shifted at every step as it is moved over the input. Note that the size of the output is less than the input and decreases even further when using a larger stride value when shifting the filter. It may also be useful to add a border of 0 values around the input to conserve the size of the output. This is known as zero padding.

Another layer fundamental to modern neural networks is the *pooling layer* that can be used to reduce the size of the inputs through mathematical operations such as averaging, maximum, etc., applied in a similar "sliding" fashion. Figure 4-5 shows an example of the output of a pooling layer using the maximum and average functions.

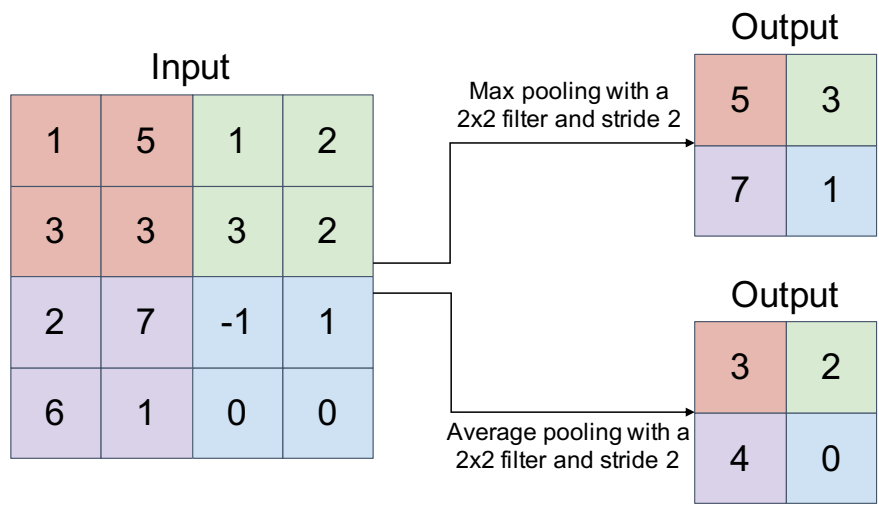

Figure 4-5 Visualization of pooling with 2x2 filters and stride 2 using maximum (top) and average (bottom) pooling.

Neural networks that use the convolutional layer are often referred to as *convolutional neural networks*. Typical network designs involve many convolutional layers, each followed by activation functions and pooling operations and finally a fully connected layer to produce the final output. For a binary classification task like the previously discussed case of tumor classification, we can use a fully connected layer comprising one neuron with a sigmoid activation to generate a valid probability of malignancy. Suppose now that we want to classify the same tumor as one of several considered histological classes using a larger fully connected layer that outputs the probability of each class. In this case, using sigmoid activations could potentially classify the tumor as all of the classes at the same time or as none of them by returning only 1s or 0s, respectively. Here, it would be more appropriate to have an activation function that converts the output into a vector of probabilities that sum to one. It follows that when one class is classified as more likely, the probability of the other classes decreases. The *softmax* activation allows us to satisfy this constraint by converting the outputs of a layer $(o_1, o_2, ..., o_n)$ into a valid probability distribution $(p_1, p_2, ..., p_n)$ using the formula:

$$p_i = \frac{e^{o_i}}{\sum_{j=1}^{n} e^{o_j}}. \qquad \text{(Eq 4-4)}$$

Taking the exponential of the class output ensures that every output is positive and the division by the sum of the exponential of every class guarantees that all the outputs sum to 1.

Figure 4-6 shows a simple design for a neural network base on AlexNet[6] that can be used to identify which surgical tools are present in an image. The HxWxC notation has been used to express the size of the output of each layer in the network, where H, W and C are the height, width and depth of the matrix, respectively. Each of the C slices of size HxW of this matrix is referred to as a *channel*.

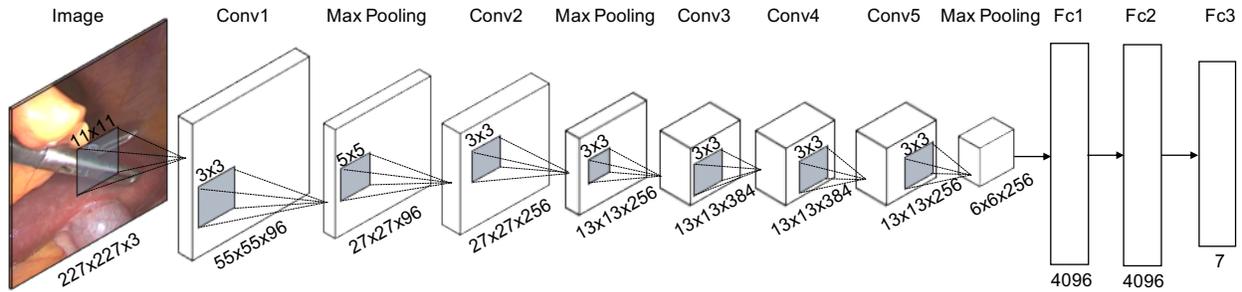

Figure 4-6 Network architecture used to generate the probability that a tool is visible in an input image.

Every pixel of the input endoscopic image can be expressed as a combination of 3 fundamental colors (red, green and blue). So, if the input image contains 227x227 pixels, it can be expressed by a matrix of 227x227x3 numeric values (one channel corresponding to each fundamental color). Note that convolutional layers are used to increase the number of channels (by using more filters) as we go deeper into the network, allowing us to extract richer semantic information as the network derives more abstract features. In contrast, the spatial size (height and width) is reduced using max pooling as we go deeper in order to reduce the number of values that must be stored and to simultaneously compress spatial information. Assuming that 7 different tools can possibly appear in any given endoscopic image, 3 fully connected layers are then used at the end of the network to aggregate these high-level representations into 7 values representing the probability that each of those tools are visible in the input.

**Training Neural Networks**

The first step to training neural networks, as with most machine learning models, is dividing the data into training and test datasets. The training dataset is a subset of the data from which the network learns; and once the network is trained, the test dataset is used to evaluate the network's ability to generalize to "unseen" data. Additionally, a third subset known as the validation dataset can be defined to test different values of manually chosen parameters, known as *hyperparameters,* that affect the training process. The validation dataset is also commonly used in the training process to assess if the model is able to properly generalize to new data. For supervised learning, it is also important to define or annotate the expected output or the *ground truth* for each data point in the dataset so that the network can learn a function that approximates the ground truth output for each respective input data point.

Unfortunately, neural networks do not come off-the-shelf with the "right" parameter values to approximate an appropriate function for every given task. Once the structure or the *architecture* of a neural network is defined in terms of the order and types of layers to be used, the network must be trained to learn effective values for its parameters to perform its intended task. This process usually begins by setting or *initializing* the parameters with random values and then defining a *loss* or *objective function $L$* which represents how far the prediction is from its ground truth value. For a *binary classification* problem like the benign versus malignant tumor case discussed previously, if the network outputs value between 0 and 1 representing the probability that a given CT scan contains a malignant tumor, a simple loss could be the *mean average error* from its ground truth, where the ground truth is defined as 1 if the tumor is malignant and 0 otherwise. Minimizing this loss through iterative updates in the parameters of the network leads, by definition, to better values of these parameters for the classification task. This process is known as *Optimization*.

Suppose the loss function $L$ is calculated based on one parameter $w$. A sample visualization of the values of $L$ (blue line) over the entire parameters space (i.e. all possible parameter choices) is show in Figure 4-7:

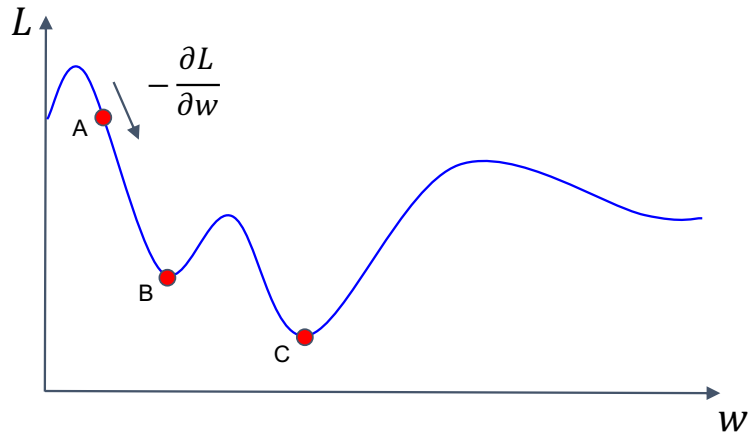

Figure 4-7 Visualization of an example loss curve. B represents a local minimum while C represents the global minimum.

The gradient $\frac{\partial L}{\partial w}$ of the loss function with respect to the parameter $w$ informs us on the direction to change $w$ to cause the largest increase in $L$. In this example where we have only one parameter, a positive value $\frac{\partial L}{\partial w}$ of implies that increasing $w$ will lead to the largest increase in $L$ and a negative value of $\frac{\partial L}{\partial w}$ tells us that decreasing $w$ would have the same effect. Logically, the negative of the gradient leads to the largest decrease in that function's value. So, for every iteration $\tau$ of the optimization process we can calculate a better value for $w$ at the next iteration $\tau + 1$ that corresponds to a lower loss. This method of minimizing the loss by following the negative direction of the gradient is an optimization technique known as *Gradient Descent*:

$$w^{\tau+1} \ = \ w^{\tau} - \lambda \frac{\partial L}{\partial w} \ .$$                              (Eq 4-5)

In Eq 4-4, the gradient informs us on which direction to change $w$ and the *learning rate $\lambda$* (where $\lambda$ > 0) is a manually chosen hyperparameter that determines how big of a step we take in that direction. If our initial random choice of $w$ was at point A (Figure 4-7), a low value of $\lambda$ may cause the network to converge to the value of $w$ at point B since it is in the direction of steepest descent for all small enough changes in $w$. B is known as a *local minimum* of the loss curve since the value of $L$ at B is less than the value of $L$ at all points within some fixed distance around B. Similarly point C is known as the *global minimum* of the curve since it corresponds to the lowest value of $L$ over the entire parameter space and so should also correspond to the target value for $w$ that we are trying to learn. Choosing a larger learning rate can help us step over the local minimum but could also potentially cause the network to step over point C and oscillate around it. Thus, a careful choice of learning rate can be a crucial factor in minimizing the loss. Finding the right value often involves some trial-and-error or alternatively, using methods such as ADAM[17], RMSProp[18] and ADAGRAD[19] that adaptively tune the learning rate during training. This learning process can be extended to any loss function irrespective of the number of parameters involved.

It is important to note that the loss and gradient mentioned in the previous paragraph is calculated for the entirety of the training dataset, often comprising thousands to millions of data points. However, loading that much data all together into the computer's memory to perform complex calculations may not always be feasible, especially when working with image data. Using *mini-batch gradient descent*, we overcome this issue by calculating the gradients and performing parameter updates on smaller subsets (known as *batches*) of the training dataset. When using batches containing only one training example each, this process is referred to as *stochastic gradient descent*. The number of times the entire training dataset passes through the network during training is denoted by the number of *epochs*. Recent advances in GPU technology, which parallelizes operations, have also made it faster to perform the matrix operations involved in this process on batches compared to providing the same number of training examples sequentially to the network.

To summarize the training procedure, for a given input we sequentially go layer by layer forward through the network until we eventually calculate the network's output. This phase of the process is aptly referred to as *forward propagation* since information flows forward through the layers of the network. Using this output, we calculate a loss representing the error in that output and then go layer by layer backwards through the network to calculate the gradient of the loss with respect to the parameters of each layer and make updates to the weights in order to decrease the loss. This is known as *backpropagation*. This entire process is repeated iteratively until a stopping criterion is met such as when our loss for the validation dataset stops decreasing or after a fixed number of iterations.

**Techniques for Deep Learning**

Training neural networks is not a trivial exercise. Over the years, several structural tricks that have emerged as standard practices in deep learning have allowed for the training of deeper networks on smaller datasets, transforming the field into a real game-changer across various domains. We will briefly describe here a few of these techniques to improve the learning process and increase a network's ability to generalize to new data.

Preprocessing: *Preprocessing* or suitably altering the dataset before passing it to the network can greatly improve the learning process in terms of stability, preventing overfitting and time to converge to a good minimum corresponding to a sufficiently low loss. Data normalization is a commonly employed preprocessing strategy where the input data is scaled down to a smaller range of values such as between 0-1. One normalization method is to *standardize* the input to be zero-centered with unit standard deviation over the training dataset. If $\mu$ and $\sigma$ are the mean and standard deviation of the inputs for all the training data, each individual input $x$ can be standardized using,

$$\bar{x} = \frac{x - \mu}{\sigma}. \qquad \text{(Eq 4-6)}$$

<u>Data Augmentation:</u> When working with small datasets, the size of the training subset can usually be inflated by adding additional data generated using modifications of the available training dataset. This is done to increase the diversity of training examples available without having to actually collect, clean and label more data. For example, when working with an image dataset, using additional data generated through small and random rotations and translations of the original dataset during training may make the model more robust to varying positions and orientations of objects in images.

<u>Regularization</u>: Large neural networks learning from small datasets tend to *overfit* to the training set and perform poorly on the test set. *Regularization* is the process of providing additional constraints to reduce overfitting. The most widely used regularization technique is *L2 regularization*, wherein an extra component for every weight parameter of the network is added to the loss function to prevent the network from learning arbitrarily large values.

$$L_{total} = L_{task} + \lambda_r \sum_{i=1}^{n} \|w_i\|^2. \qquad \text{(Eq 4-7)}$$

where $n$ is the total number of weights in the network and the *regularization strength* $\lambda_r$ is a manually set hyperparameter that needs to be carefully chosen. This effectively prevents the network from learning overly complex functions that fit perfectly to the training data but do not generalize well to new data. Regularizing the bias parameters usually has much less of an impact towards preventing overfitting compared to weight regularization and so is not as common.

As neural networks are trained, the contribution of certain pathways of neurons in the network may gradually become more and more irrelevant or crucial towards making correct predictions on the training dataset. Dropout is the technique of excluding random subsets of neurons from training at every iteration to encourage these neurons to learn representations less reliant on specific pathways. This effectively forces the network to learn smaller weights corresponding to multiple paths rather than large weights along specific ones. Dropout can thus provide a similar regularizing effect as L2 regularization.

<u>Transfer learning:</u> In practice it may be very difficult to learn optimal parameter values from scratch (i.e. randomly initializing parameters values), especially when working with large networks and small datasets. One strategy is to pre-train the network on an alternate but similar task and/or dataset and then optimize or finetune a subset of the network's parameters on the target dataset. This is known as *Transfer Learning*. Sometimes, in order to reduce the number of parameters that have to be learned, only the last layers are finetuned on the dataset of interest. In such cases, parameter values for the lower layers are not retrained since they are thought to correspond to more generic features like edges which apply to varying datasets. This approach is commonly used in surgical research where networks that have been pre-trained on the ImageNet database are subsequently fine-tuned on a more specific surgical dataset such as research group CAMMA's

cholec80 dataset of laparoscopic cholecystectomy or MGH SAIIL's SleeveNet database of laparoscopic sleeve gastrectomy.

## Deep Learning Tasks

A good starting point when deciding network architectures to solve a problem may be to look for networks that solve similar tasks and adapt them to the specific needs and data if necessary. This section will briefly describe some broad types of problems that deep learning is particularly efficient at solving and some of the involved network architectures. Most deep learning tasks fit into one of three categories based on the type of output expected from the network: classification, detection and semantic segmentation. We would like to stress that these categories are neither exhaustive nor unrelated.

## Classification

Classification refers to the task of categorizing a given input into two or more possible classes (see Figure 4-9a). For example, a classification network could be used to identify a lesion on a dermoscopic image as a benign or malignant, or to classify an ECG segment as normal or as showing an arrythmia. Note that lower dimensional convolutional filters are used when operating on lower dimensional inputs such as ECG segments, which consist of voltage values. Some network architectures frequently used for deep learning tasks involving classification on images are Resnet[20] and VGG net[21] (Figure 4-8). Most classification architectures involve a convolutional network ending with a fully connected layer with one neuron for every considered class. These neurons are generally used to output the probability of the input belonging to each class.

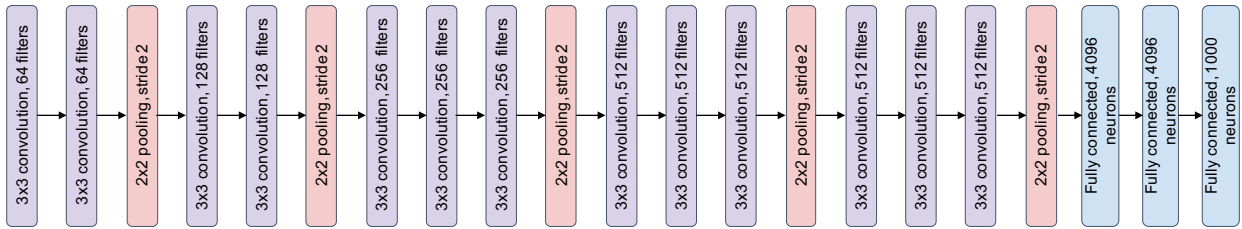

Figure 4-8 Visualization of the VGGnet architecture used for classification. Here 3x3 convolution, 64 implies that 64 3x3 convolutional filters are used to operate on that layers input; the fully connected layers have 4096 outputs or neurons. Each time the spatial size is halved using a pooling layer, a convolutional layer is used to double the number of channels.

## Detection

Detection, like classification involves identifying the category of an object of interest in the input but additionally localizing its position spatially, as illustrated in Figure 4-9b. For image data, the position of a target object could be described by coordinates of the corners of a bounding box around the object predicted by the network. For every object of interest in the input, ground truth annotations (in a supervised setting) describe both the class of the object as well as sufficient information to completely describe the position of the object. An example of where neural networks have been used to achieve strong performance in a detection task is the identification and localization of tools in endoscopic images[22].

## Semantic segmentation

Semantic segmentation is the task of classifying every pixel of an image into a particular category (Figure 4-9c). As an example, segmentation of the cerebrovascular network in a magnetic resonance angiography[23] could be a useful tool for clinicians and radiologists to interpret those images because of the large inter-patient variability of anatomy in such scans. Some network architectures that are commonly used by the computer vision community to solve segmentation problems are Deeplabv3+[24] and HRnet[25]. Unet[14] is another popular architecture that has shown promising results for semantic segmentation on medical images.

Unlike object detection and classification, semantic segmentation involves generating a much larger output representing predictions corresponding to every pixel of the input. Maintaining the input feature size at all layers of a very deep neural network is computationally intensive both in terms of speed and memory. To solve this issue, segmentation networks such as Deeplabv3+ and Unet use an encoder-decoder framework. These networks first use an encoder path to extract rich semantic information at reduced dimensions and then a decoder path to recover sharp boundaries while restoring the original dimensionality. An example of a naive decoder would be a simple resizing operation.

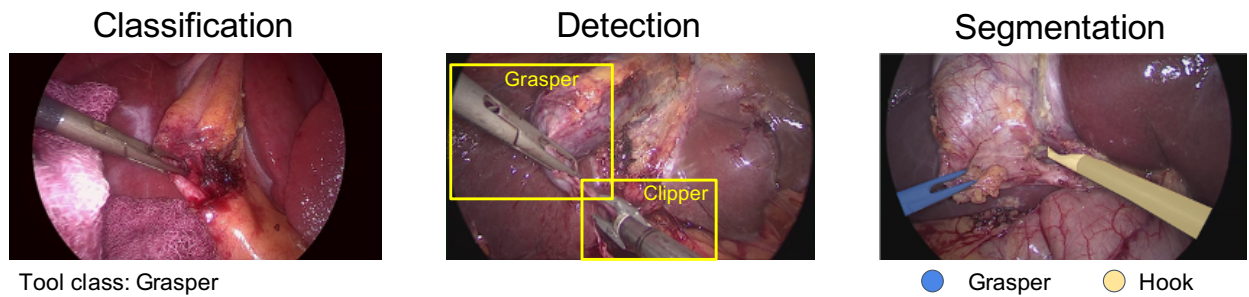

Figure 4-9 Visualization of network outputs for (a) tool classification, (b) tool detection and (c) tool segmentation performed on images extracted from a laparoscopic cholecystectomy video.

## Synthetic data generation

Synthetic data generation is the task of creating new data mimicking the characteristics of a given training dataset. This can be achieved using generative networks, a kind of network constructed to learn the statistical distribution of a dataset. Given the dearth of large and well-labelled medical datasets, data generation approaches have proven useful to augment small datasets as well as generate entirely artificial ones. Such artificially augmented datasets have been successfully employed to improve performance for lesion classification[26]. Similar approaches have also been used to modify radiological scans, with one article[27] even demonstrating a vulnerability of hospitals to CT scan attacks where a generative network is used to inject or remove evidence of cancer, potentially leading to misdiagnosis.

One commonly employed framework for data generation is the *generative adversarial network* (GAN), as depicted in Figure 4-10. In this framework, a model is trained to synthesize images resembling examples from a given training dataset using two neural networks. The first network, or generator, tries to synthesize an output resembling the training data while a second network, or discriminator, competes to identify whether the generated example was a fake. During training, as the discriminator gets better at discerning fake examples, the generator becomes more efficient at generating realistic fakes.

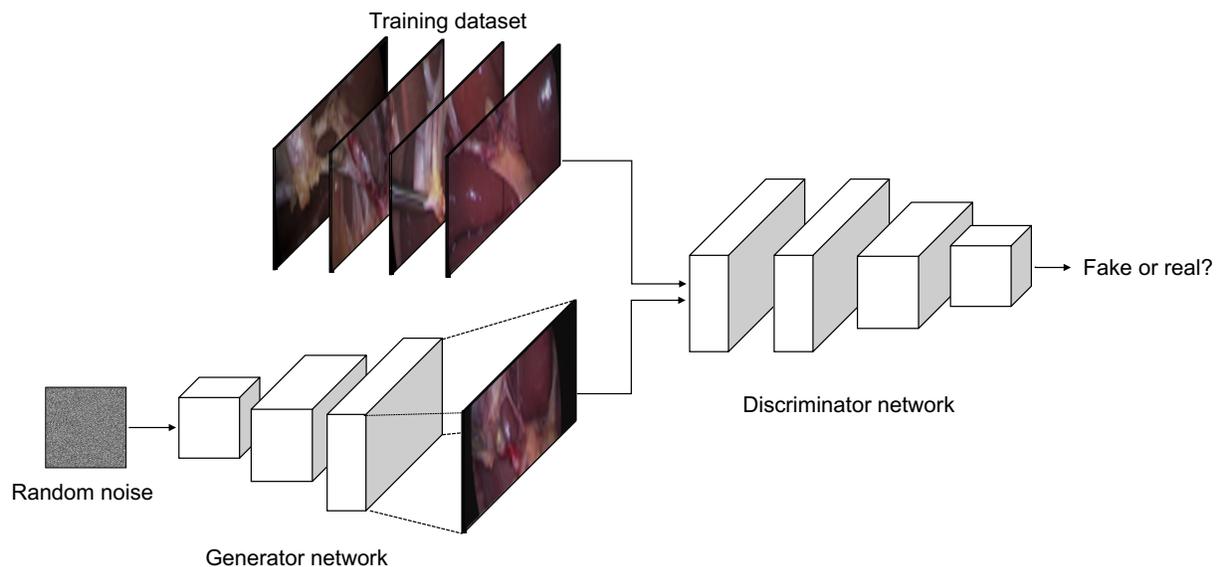

Figure 4-10 An example of a GAN being used to generate realistic laparoscopic images.

Another useful approach for generative modelling is to use *autoencoder* networks that first encode an input into a latent representation of lower dimensionality and then decode that representation to reconstruct the original image. Due to the dimensionality reduction in the first stage, these representations learn to encode only the most relevant features of the input needed for its reconstruction and to ignore noise. If encodings from real images are replaced by realistic guesses of latent representations, the decoding stage of the network can then be used to generate new examples. *Variational autoencoders* (VAE) leverage this property by constraining the network, through an additional loss term, to learn representations for different inputs that are sufficiently close numerically while still remaining distinct. This effectively forces the latent representations to represent common features whose variations are sufficiently descriptive to reconstruct realistic variations of the original input. An added benefit is that these representations are often easier to interpret. For example, when training a VAE to generate laparoscopic images, the variables that constitute the latent representation could correspond to size, orientation and position of different anatomical structures in the image.

**Recognition with temporal models**

Imagine trying to identify the type of surgical procedure being performed from just a single still endoscopic image versus having a sequence of frames or a video. For tasks involving sequential data, it may be useful and sometimes necessary to add a component to the network architecture that "remembers" information derived from previous inputs in the sequence. Recurrent Neural Networks (RNN) add this concept of memory to the network by adding a loop between layers that allows information to persist in the network from previous steps in the sequence through a *hidden state* (Figure 4-11).

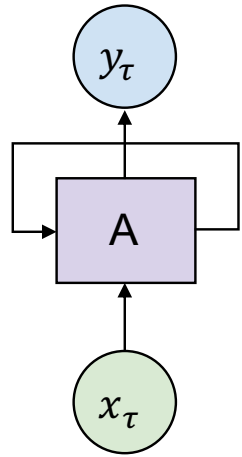

Figure 4-11 Recurrent neural network receiving an input at time step $\tau$ and generating an output.

At every step $\tau$, the RNN receives both the input $x_\tau$ for that time step as well as a learned hidden state $h$ from step $\tau - 1$. An unrolled depiction of the RNN is shown in Figure 4-12. An initial assumption must be made for what will be passed to the network as the hidden state for the first step $\tau = 1$.

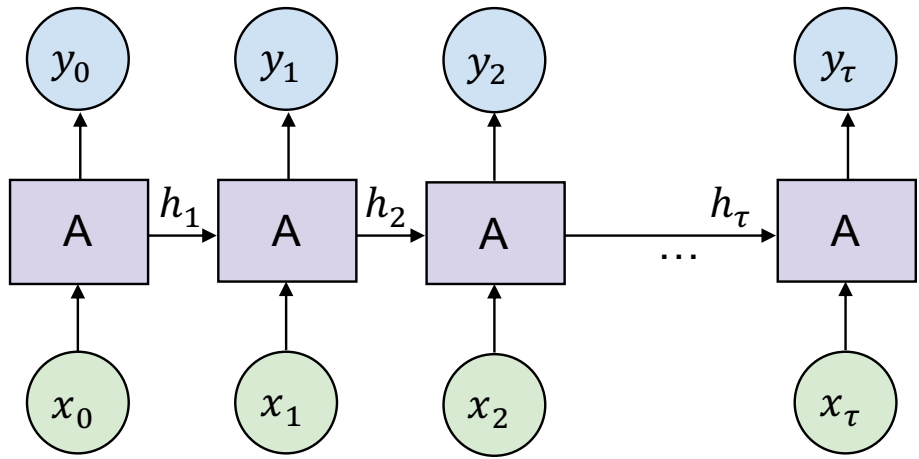

Figure 4-12 Unrolled depiction of a RNN receiving inputs at different timesteps.

When backpropagation is performed on such a network, the gradients are calculated on the unrolled version of the RNN. This is known as *Backpropagation Through Time (BPTT).* Without going too deep into the mathematics of the gradient computation, this could lead to a numerical instability in the gradients that causes it to grow or shrink to arbitrarily large or small values respectively. This predicament is known as the "vanishing or exploding gradients" problem and could severely hinder the learning process. Another major issue introduced by this kind of network is that a sequence of irrelevant data caused by some event such as an obstructed camera view may cause useless information to be propagated through the hidden state even though relevant inputs were available to the network at earlier steps. Two variants of RNNs that address these problems are the Long-Short Term Memory (LSTM)[28] model and the Gated Recurrent Unit (GRU)[29]. These models introduce separate neurons that control what information is stored and discarded for each step in the sequence of inputs. This helps filter out irrelevant information and emphasize important information.

Often, convolutional neural networks are first used to extract meaningful features that are then fed as input to an RNN that models temporal relationships between these features at different steps in the sequence. For instance, these kinds of networks have been used for early prediction of sepsis[30] based on clinical time-series data such as blood pressure, body temperature, blood glucose, etc. Some other surgical applications where LSTM based models have demonstrated great success are surgical skill evaluation[31] and surgical activity recognition[32].

## DEEP LEARNING IN SURGERY

At this point, the reader should be familiar with the terminology necessary to read and think about deep learning. Here we would like to examine the implementation of a deep learning pipeline to guide the reader through the application of concepts introduced earlier. These concepts can further be put in practice in the hands-on lab provided with the chapter, available at https://github.com/CAMMA-public/ai4surgery. Surgical phase and tool detection models[15], two early examples of deep learning for surgical applications, will serve as examples. Detecting workflow and tool usage (Fig 4-13) is a necessary step to understand the surgical context, an essential milestone towards smart support or autonomous assistance systems[33, 34].

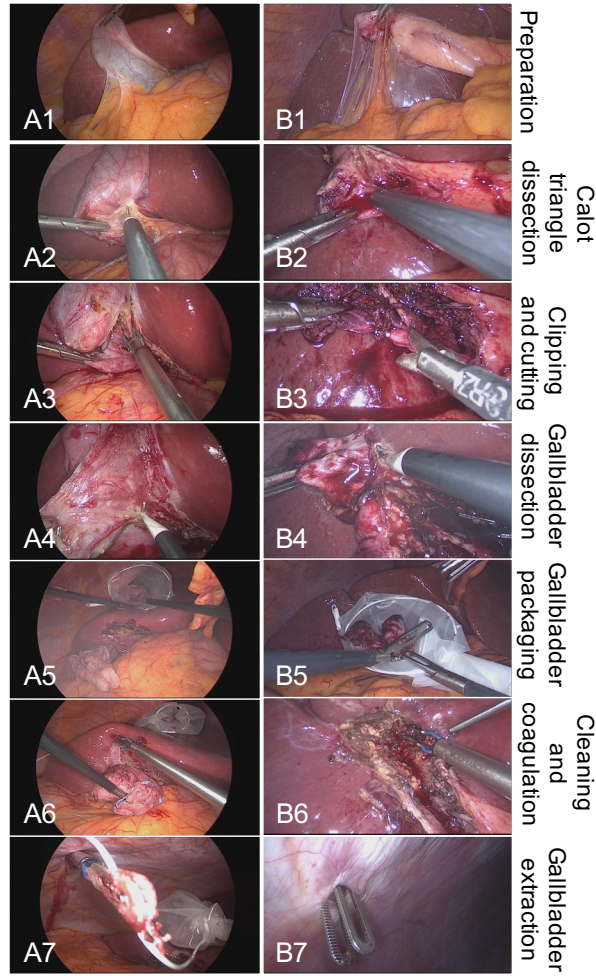

Figure 4-13 Endoscopic images from two laparoscopic cholecystectomy procedures (A and B) representing the 7 phases defined in the Cholec80 dataset[15]. The images contain the surgical tools typically used during cholecystectomy. Images taken from procedure A have a black frame around the endoscopic scene while the others have none, showing the need for the preprocessing of endoscopic images.

## Data and Annotation

Choosing appropriate input data for a model to learn a given task is of fundamental importance.

In most cases, a simple and effective way to check whether data is sufficiently informative is to ask an expert to look at it and perform the same task the model will be trained for. In our examples, given an endoscopic picture one can easily say if a tool is present but cannot easily infer the phase of a procedure from it. Videos are more suitable than pictures for tasks like workflow (i.e. phases and activity) detection for which knowing the temporal context intuitively helps. Unfortunately, there is no straightforward way to estimate how much data is needed other than gauging the variability in input data and evaluating model performance with datasets of different sizes. In classification tasks like tool or phase detection, estimating inter-class and intra-class similarities or variability can help understand the complexity of the problem. For instance, distinguishing a surgical hook from a scissor is clearly easier than differentiating a crocodile grasper from a Maryland grasper. Since the two types of graspers (i.e. the two classes) are similar, an observer (or an algorithm) would need more experience, i.e. more data, to learn to differentiate between them. In the case of surgical workflow analysis, the initial phase of a cholecystectomy could be either very short and straightforward or require extensive adhesiolysis. That is saying, the same phase of a procedure can look very different, i.e. there is a large intra-class variability, so to learn how to correctly classify surgical phases this variability needs to be well represented in the dataset. Informativeness of input data, inter-class and intra-class variability also influence the consistency of annotations, a fundamental factor influencing model performance. In supervised learning, models are trained to approximate a function to go from the input data to the annotated information. It follows that if annotations are not consistent the model will compute a loss on incongruent data, deteriorating performance. In our example, persons with no medical education could correctly annotate tool presence or absence in images. The same cannot be said for annotating surgical phases since those require surgical understanding and a shared definition of what exactly describes and delineates phases.

## Implementation

We briefly present below how to implement deep learning on surgical images. More details on data acquisition, storage, and annotation of surgical videos can be found in Chapter 16 - *Practical Consideration in Utilization of Computer Vision.* First, data is analyzed and pre-processed. For phase and tool detection, preprocessing may include resizing, cropping (see Figure 4-12) and padding input images and/or excluding data annotated with inconsistent labels. Then, the database is split into training, validation and test sets. It is good scientific practice to keep these sets as independent from each other as possible since the network may develop biases (not to be confused with the network parameter) based on visual patterns not relevant to the network's task. Furthermore, when reporting results of deep learning experiments, it is important to carefully describe the dataset, its splits and applied pre-processing since this may affect interpretation of outcomes.

At this point, one or more deep-learning architectures showing good performance on similar tasks or data are identified. Classifiers like ResNet, VGG or NASNet may work for phase and tool detection in images. However, as seen earlier, the temporal information captured by the video can be extremely relevant, at least for phase detection. Adding an RNN like a LSTM to model the temporal relationships could thus boost performance.

Before starting the learning process the model parameters need to be initialized. Weights and biases can be initialized randomly or, in transfer learning, taken from the same model pre-trained on a large dataset. The latter is often the case in surgical data science given the limited size of annotated datasets. At this point we are ready to start training the model.

In the case of a classification task such as phase and tool detection, the initialized model will output class scores reflecting the computed probability of each surgical tool or phase being represented in the input. Various loss functions can then be computed to quantify the error between class probabilities and ground truth. For instance, building on the surgical intuition that certain tools are more frequently used in certain surgical phases, as one can appreciate in Figure 4-12, a model for tool detection can compute a loss on the probability of tool and phase co-occurrence[35]. Once a suitable loss function is defined, gradients are calculated, and neurons parameters are updated. This process is repeated for a number of iterations, usually until results plateau or after a predefined number of iterations. Before testing on unseen data, hyperparameters like learning rate and batch size are tuned on the validation dataset.

The model is then finally run on the test dataset and results are reported according to performance metrics chosen on the basis of the nature of the task and the composition of the dataset. These results are usually compared against one or more baselines. When aiming at improving performance of an already tackled problem like phase and tool detection, a good baseline for comparison is usually obtained by re-implementing related state-of-the-art architectures on the data of interest. However, when tackling a completely novel problem the baseline is often represented by the results of a naive or random model - the equivalent of flipping a coin for a binary classification task. Further, the contribution of each part of the proposed model to the outcome can be examined through an *ablation study* consisting in sequentially removing components of the model to see how performance changes. Finally, performance of deep learning models should be examined in the light of the intended clinical use. For instance, not detecting a grasper in an endoscopic image will have little to no impact while a false negative in a screening mammography can have dramatic consequences, and so, metrics must be chosen that reflect these clinical priorities.

## CONCLUSIONS AND PERSPECTIVES

Even though biologically inspired artificial neural networks have been around for a while, deep networks for real world applications have only recently appeared. This poses new fascinating questions to researchers and stakeholders, especially when looking at applications in highly regulated fields like medicine and surgery. We will now briefly touch some of the questions researchers are trying to address to translate the potential of neural networks to surgery.

Data privacy, security and ownerships are issues common to most data-driven approaches, especially when treating highly sensitive health data. It is beyond the scope of the present chapter to describe all possible methods to guarantee data privacy and security; however, it is important to know that deep learning can help in this regard. For instance, face detection models have been used to automatically hide identity of people recorded in the OR[36] and models capable of analyzing data acquired in an identity-preserving format such as recording at extremely low-resolutions[37] or with depth and thermal sensors[38] have shown promising results. Further, distributed learning approaches could allow training and validating algorithms on varied data generated in multiple institutions without the need to centralize, hence move and share, any data.

On top of the mentioned issues with clinical data, annotating raw data with surgical knowledge requires rare and expensive physician-time. Hence an important bottleneck slowing the development of deep learning models in surgery is the scarcity of well annotated datasets. Successes with deep networks in other domains are in great part fueled by big datasets like ImageNet, a dataset containing about 14 million images annotated with more than 20000 classes of objects[7]. In surgery we are far from having such a vast and well annotated database, hence methods able to learn on less data and with less supervision are the focus of many research groups. For instance, methods for tool detection[39], tracking[22], segmentation[40] and phase detection[41] requiring very little annotations have been proposed.

Furthermore, understanding how and why a model returns a certain prediction is a priority, especially in the medical field due to medico-legal considerations. Unfortunately, deep learning approaches have been criticized as being black boxes since their internal logic is hard to explain. However, recent works such on *saliency maps[42]* and *class activation maps[43]*, methods that determine which parts of the input affect the output, have shown progress in improving the interpretability of deep-learning models. For example, Al Hajj and colleagues found that their deep model was "focusing" on the deformation of the anterior segment of the eye to infer if surgical tools were in contact with the eye in cataract procedures[44].

Finally, to impact surgical care, models will have to communicate or be integrated within ORs. Real-time OR-embedded vision applications could be possible with lightweight and efficient architectures such as mobile-net[45] and efficient-net[46] that uses a computationally superior variant of convolution operation known as *depth-wise separable convolution*. Another strategy to improve model's run time is to use the *mixed-precision* operation. Latest versions of NVIDIA GPUs, for example the NVIDIA-V100, include *tensor-cores* that can perform a combination of low and high precision operations, i.e. approximate numbers to different exactness levels. Mixed-precision computations are much more efficient than high-precision computation but are often considered much less accurate. This is disputed by recent studies[47] showing that, using better training techniques, mixed-precision computation can achieve performance comparable to high-precision computations.

In conclusion, despite the novelty of deep learning applications in surgery and the numerous open challenges to address, the ability of neural networks and deep learning to extract meaningful information from the vast amount of raw data we continuously generate is likely to redefine surgical practices. It is now the ideal moment to join the discussion, team up with stakeholders and help shape the impact of deep learning on patients and surgical practices.

**HANDS-ON DEEP LEARNING LAB**

A hands-on practical on deep learning for surgical data is available online as companion to this chapter. The interested reader can directly experiment with the methods presented in this chapter by following this link:

https://github.com/CAMMA-public/ai4surgery.